%% file: main.tex
\documentclass[twocolumn]{article}
\usepackage{spconf,amsmath,graphicx}

\usepackage{times}
\usepackage{epsfig}
\usepackage{graphicx}
\usepackage{amsmath}
\usepackage{bbm, dsfont}
\usepackage{amssymb}
\usepackage{booktabs}
\usepackage[dvipsnames, table]{xcolor}
\definecolor{Gray}{gray}{0.85}
\usepackage{multirow}
\usepackage{pifont}
\usepackage{algorithm}
\usepackage{algpseudocode}
\usepackage{comment}
\usepackage{hyperref}
\usepackage{subcaption}
\usepackage{caption}
\usepackage{tikz}
\usepackage{comment}
\usepackage{amsmath,amssymb} 
\usepackage{color}
\usepackage{amsthm}

\def\*#1{\mathbf{#1}}

\title{Fighting Over-fitting with Quantization for Learning Deep Neural Networks on Noisy Labels}
\name{Gauthier Tallec $^{1,*}$, Edouard Yvinec$^{1,2,*}$, Arnaud Dapogny$^2$, Kevin Bailly$^{1,2}$ \thanks{$*$ stands for equal contribution. This work has been supported by the french National Association for Research and Technology (ANRT), the company Datakalab (CIFRE convention C20/1396) and by the French National Agency (ANR)  (FacIL, project ANR-17-CE33-0002). This work was granted access to the HPC resources of IDRIS under the allocation 2022-AD011013384 made by GENCI.}}
\address{Sorbonne Université$^1$, CNRS, ISIR, f-75005, 4 Place Jussieu 75005 Paris, France \and Datakalab$^2$, 114 boulevard Malesherbes, 75017 Paris, France}
\begin{document}
%
\maketitle
\begin{abstract}
The rising performance of deep neural networks is often empirically attributed to an increase in the available computational power, which allows complex models to be trained upon large amounts of annotated data. However, increased model complexity leads to costly deployment of modern neural networks, while gathering such amounts of data requires huge costs to avoid label noise. In this work, we study the ability of compression methods to tackle both of these problems at once. We hypothesize that quantization-aware training, by restricting the expressivity of neural networks, behaves as a regularization. Thus, it may help fighting overfitting on noisy data while also allowing for the compression of the model at inference. 
We first validate this claim on a controlled test with manually introduced label noise. Furthermore, we also test the proposed method on Facial Action Unit detection, where labels are typically noisy due to the subtlety of the task. In all cases, our results suggests that quantization significantly improve the results compared with existing baselines, regularization as well as other compression methods. 
\end{abstract}

\section{Introduction}

\begin{figure}
    \centering
    \includegraphics[width=8.6cm]{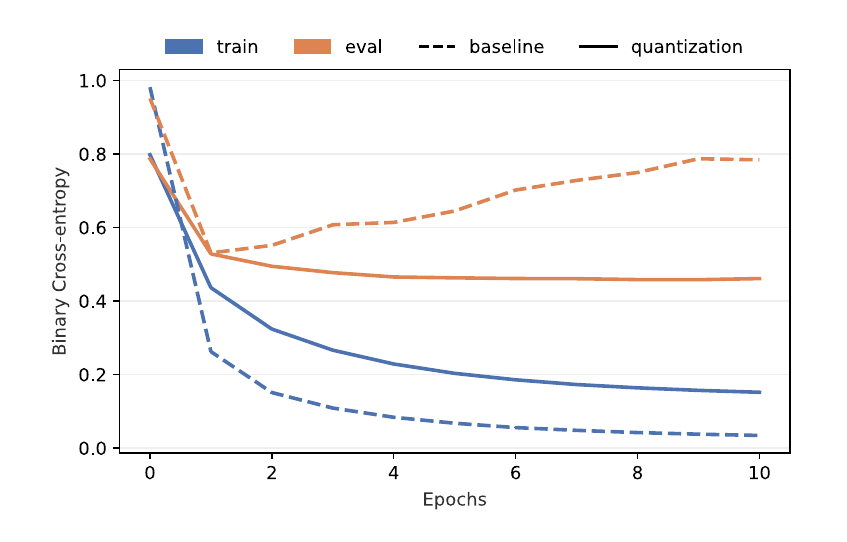}
    \caption{Training (blue) and evaluation (orange) loss between baseline (dashed) training and quantization aware (plain lines) training on BP4D. While the baseline network quickly overfits (as indicated by the eval error rising while the training error decreases) due to noisy labels, the quantized network noticeably avoids this pitfall.}
    \label{fig:bp4d_quantvbaseline}
\end{figure}
In the recent years, a wide range of computer vision tasks have benefited from the public release of large scale datasets, including but not limited to image classification \cite{imagenet_cvpr09}, object detection \cite{lin2014microsoft}, image segmentation \cite{zhou2017scene} and face analysis \cite{wood2021fake}. These datasets, in combination with the growing availability of the resources required to train over-parametrized neural networks, have enabled computer vision solutions to achieve remarkable performance \cite{carion2020end,touvron2021going,tallec2022multi}. However, this trend is undermined by two major factors: difficulty of deployment \cite{jacob2018quantization,yvinec2023powerquant} and ill annotated datasets \cite{tommasi2017deeper,kuznetsova2020open}. While the former can be tackled by using compression techniques \cite{jacob2018quantization,yvinec2022singe,yvinec2023powerquant}, the latter is more challenging to address. The main reason behind this difficulty to detect and correct the wrong annotations being the sheer size of these sets of millions of examples as well as the subtlety of the task (\textit{i.e.} low inter-agreement among experts).

Due to their ability to approximate any continuous function \cite{funahashi1989approximate}, over-parametrized neural networks can over-fit on such noisy datasets which hinders their ability to generalize to properly annotated sets. This issue falls in the broader problem of overfitting. Regardless of the quality of the annotations, their exists a wide range of methods to avoid overfitting. The most notable ones being early stopping \cite{prechelt1998early}, weight decay \cite{loshchilov2018fixing}, label smoothing \cite{szegedy2016rethinking} and dropout \cite{srivastava2014dropout}. 

More generally, the techniques that limit the expressive power of neural networks are well known for their ability to reduce overfitting of over-parameterized networks. Such reduction of expressivity can be achieved with neural network compression methods.
For instance, a simple manner to reduce expressivity of neural networks is pruning and consists in removing neurons and feature maps during or prior to training. In \cite{gerum2020sparsity}, pruning prevents overfitting in the case of classification, during training, by progressively discarding the least important features. Bramer \cite{bramer2002pre}, performed a similar study on classification trees with pruning at initialization. Another effective compression approach is quantization, which maps floating point operations to low-bit fixed point operations and as such reduces the expressivity of the model. Such techniques have been applied to enhance robustness to adversarial attacks \cite{yoon2018stochastic} which leads to an improved generalization and reduced overfitting.
In \cite{xu2018quantization}, the authors apply quantization to improve suggestive annotations \cite{yang2017suggestive}, which extracts a representative and small-sized balanced training dataset based on uncertainty metrics. This method reduces overfitting on small training samples.

To the best of our knowledge, the ability of such techniques to fight overfitting on noisy data has never been tested. Instead, prior works focused on the effects of compression on the learning procedure and resulting models which were assumed to be trained on perfectly labeled data. To complement the existing results, we hypothesize that \textit{quantization-aware training, by restricting the expressivity of neural networks, is a very effective method for limiting overfitting on noisy annotated data}, as can be observed in Fig. \ref{fig:bp4d_quantvbaseline}. To test this hypothesis, we conducted a comparison to the most commonly used regularization techniques and also pruning.

\section{Fighting overfitting in Deep Neural Networks}
\subsection{Regularization Techniques}

Regularization techniques are defined as any technique that helps improve the model generalization \cite{kukavcka2017regularization}. Usually, these methods focus on improving accuracy on a defined test set and do not change the computations required at inference. The most commonly used ones are early stopping \cite{prechelt1998early}, weight decay \cite{loshchilov2018fixing}, dropout \cite{srivastava2014dropout} and label smoothing \cite{szegedy2016rethinking}. For what follows, let's consider a neural network $F$ with $L$ layers $f_l$ and weights $W_l$.

\noindent\textbf{Early Stopping:} As we train $F$, the accuracy on the training and validation data increases. When overfitting occurs (Fig. \ref{fig:bp4d_quantvbaseline} a.), the accuracy on the validation and the test sets starts to decrease while it continues to increase on the training set. Early stopping consists in using a validation set in order to find when to stop training the model.

\noindent\textbf{Weight Decay:} Weight decay fights overfitting by introducing a prior on the scale of all the weight values in $F$. Concretely, it consists in the addition of an L2 penalization term to the training loss $\mathcal{L}_{w} = \alpha_{w} \sum_{l=1}^{L} \|W_l\|^{2}_2$ where $1 / \alpha_{w}$ is proportional to the scale imposed on the weights.

\noindent\textbf{Dropout:} Dropout curbs overfitting by randomly masking parts of $F$ in order to learn to predict with a sub-network, avoiding co-adaptation of the weights. Formally, during training, for each example, each scalar weight values is set to $0$ with probability $p$. In test, all weights are multiplied by $p$ to account for their frequency of presence in train.

\noindent\textbf{Label Smoothing:} Label smoothing reduces overfitting by preventing neural network over-confidence. To do so it modifies the ground truth label as follows : 
\begin{equation}
    \*y_{\alpha_{s}} = (1 - \alpha_{s}) \*y + \alpha_{s} \frac{1}{C},
\end{equation}
where $\alpha_{s}$ controls the smoothing intensity and $C$ is the number of classes in the classification problem. For multi-task binary classification (e.g. action unit (AU) detection), label smoothing is applied label-wise with $C=2$ for presence and absence of a given label.

Note that all the aforementioned methods do not change the inference runtime of the network, hence not addressing the efficiency problem. For this reason, we propose to evaluate the ability of compression techniques for regularization.

\subsection{Compression techniques}
\noindent\textbf{Pruning:} Let's assume that $F$ is pre-trained. Then for each $f_l$, we perform standard magnitude-based structured pruning \cite{molchanov2016pruning} over the weight tensors $W_l$ by removing the neurons with highest $L^1$ norm. The method follows from the intuition that smaller weights induce smaller activation which themselves contribute less to the decision making. Consequently, as the expressivity of the model is reduced, it is less likely to overfit.

\noindent\textbf{Quantization:} The standard quantization operator, in $b$ bits, is defined by $\text{quantized}(X) = \left\lfloor X \frac{2^{b-1}-1}{\lambda_X} \right\rceil$ where $\left\lfloor \cdot \right\rceil$ is the rounding operation and $\lambda_X$ is scaling parameter specific to $X$ which ensures that $X$ support is correctly mapped to $[- (2^{b-1}-1) ; 2^{b-1}-1]$. It is common to have scalar values for $\lambda_X$ when quantizing activations (\textit{i.e.} layer inputs) and vector values for weight tensors (per-channel quantization). The activation scales are estimated per-batch during training $\lambda_X = \max\{|X|\}$ and the inference value is updated using an exponential moving average. On the other hand, weight scales are always computed as the maximum per-channel of $|W_l|$.
When optimizing the weight values $W$, the rounding operator introduces zero gradients almost everywhere which is problematic for gradient-based optimization. To circumvent this limitation, straight through estimation \cite{bengio2013estimating} is an efficient solution as the gradient operator associated to the quantization process is replaced by the identity function. In this method, the batch-normalization layers are removed from the network architecture. 

Consequently, we argue that, by limiting the representative power of $F$, quantization and pruning act as regularization during training while also significantly improving speed at inference time. In our experiments, we show the ability of quantization to out-perform all the aforementioned regularization methods when applied on noisy training datasets.

\section{Experiments}

\subsection{Datasets}
We test our hypothesis on two set-ups: first single task classification on Cifar10 \cite{krizhevsky2009learning} where we manually introduce noise on the labels, second on BP4D \cite{zhang2014bp4d}, a multi-task AU detection dataset, where the annotations are expected to be slightly imperfect. 

\noindent \textbf{Cifar10} is a classification dataset comprising 50,000 training images and 10,000 test images, all annotated in 10 classes. We use this dataset to simulate the robustness to bad training annotations by sampling $s\%$ training examples and randomly re-annotating them, \textit{i.e.} we know that $s\%$ of the modified training set have a random label but discard that information during training. Note that the test set remains unchanged in all cases. 
To tackle this task, we train a ResNet 20 \cite{he2016deep} for 200 epochs using Adam \cite{kingma2014adam} optimizer.

\noindent \textbf{BP4D} is a dataset for facial AU detection and comprises about $140k$ images featuring $41$ people. Each image is annotated with the presence of $12$ AU. For performance evaluation, we follow related work strategy from \cite{shao2018deep} and use the vanilla architecture from \cite{tallec2022multi}. For stability concerns, the evaluation performance are averaged over 5 runs.

\subsection{Implementation Details}
Each of the aforementioned regularization and compression method requires hyper-parameters. For the vanilla regularization techniques, we use widely adopted hyper-parameters: weight decay $w_d = 0.01$, dropout proportion $p=0.1$, and smoothing parameter $\alpha_{s} = 0.1$. On BP4D, we systematically apply early stopping as described in \cite{tallec2022multi}.

To achieve good performance with quantization on multi tasking, we adapted straight-through estimator by keeping batch-normalization layers, in order to learn the input scaling factors and consequently be robust to strong discrepancies between tasks. Formally, we keep the normalization process per-task (per-AU). This change was required to get stable results across several runs. 

\begin{figure}
\centering
\includegraphics[width = \linewidth]{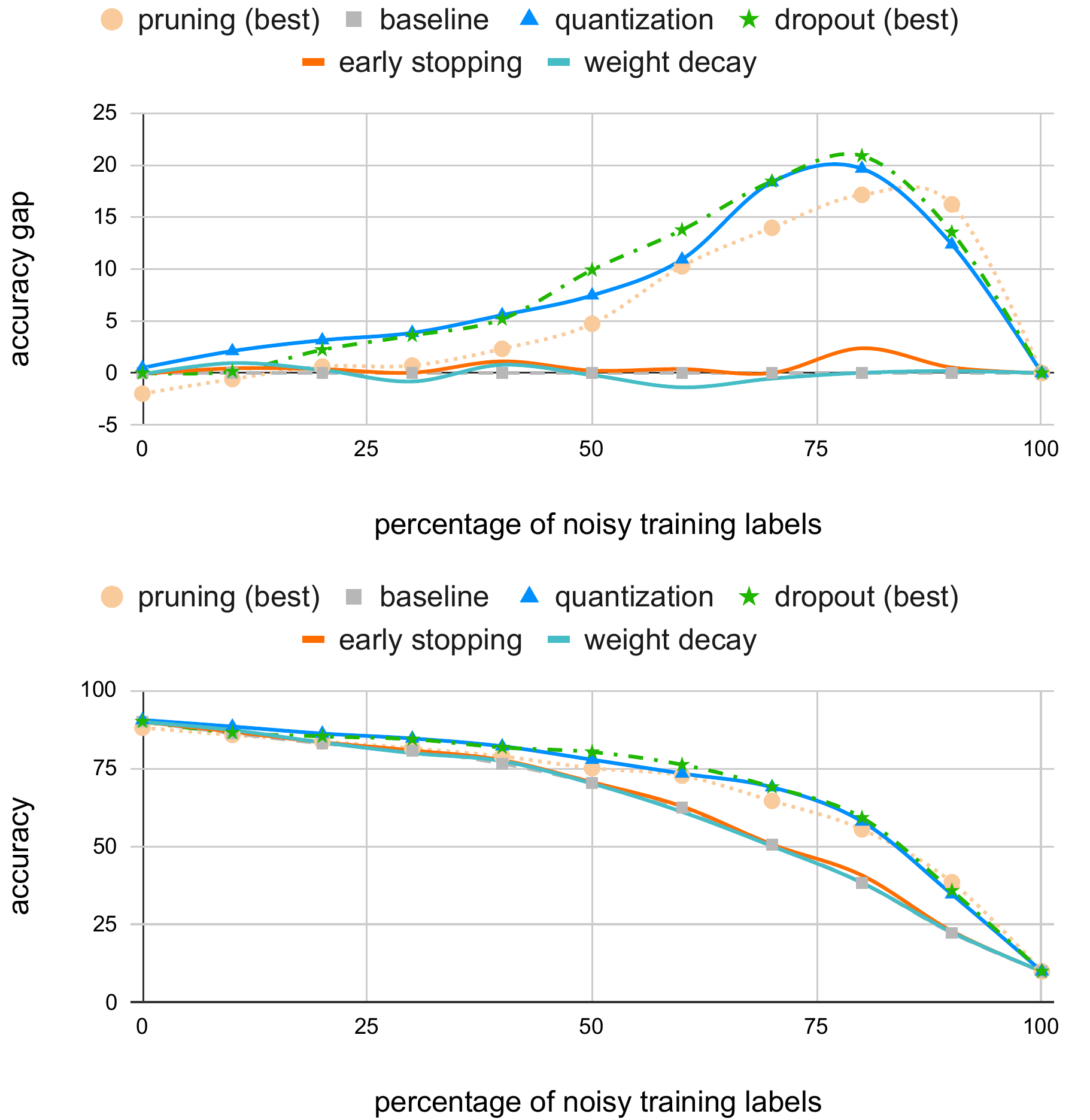}
\caption{On Cifar10: (a) Accuracy gain with respect to the baseline model (no regularization applied) as a function of $s$. (b) Accuracy as a function of the percentage $s$ of training examples randomly re-annotated.}
\label{fig:cifar10_performance}
\end{figure}
\begin{table*}
    \centering
	\input{tables/bp4d/regularization_comparaison.tex}
    \caption{Comparison between Quantization, Pruning and other vanilla regularization methods on BP4D \label{tab:regul_compare}}
\end{table*}

\begin{figure}[!t]
\centering
\includegraphics[width = \linewidth]{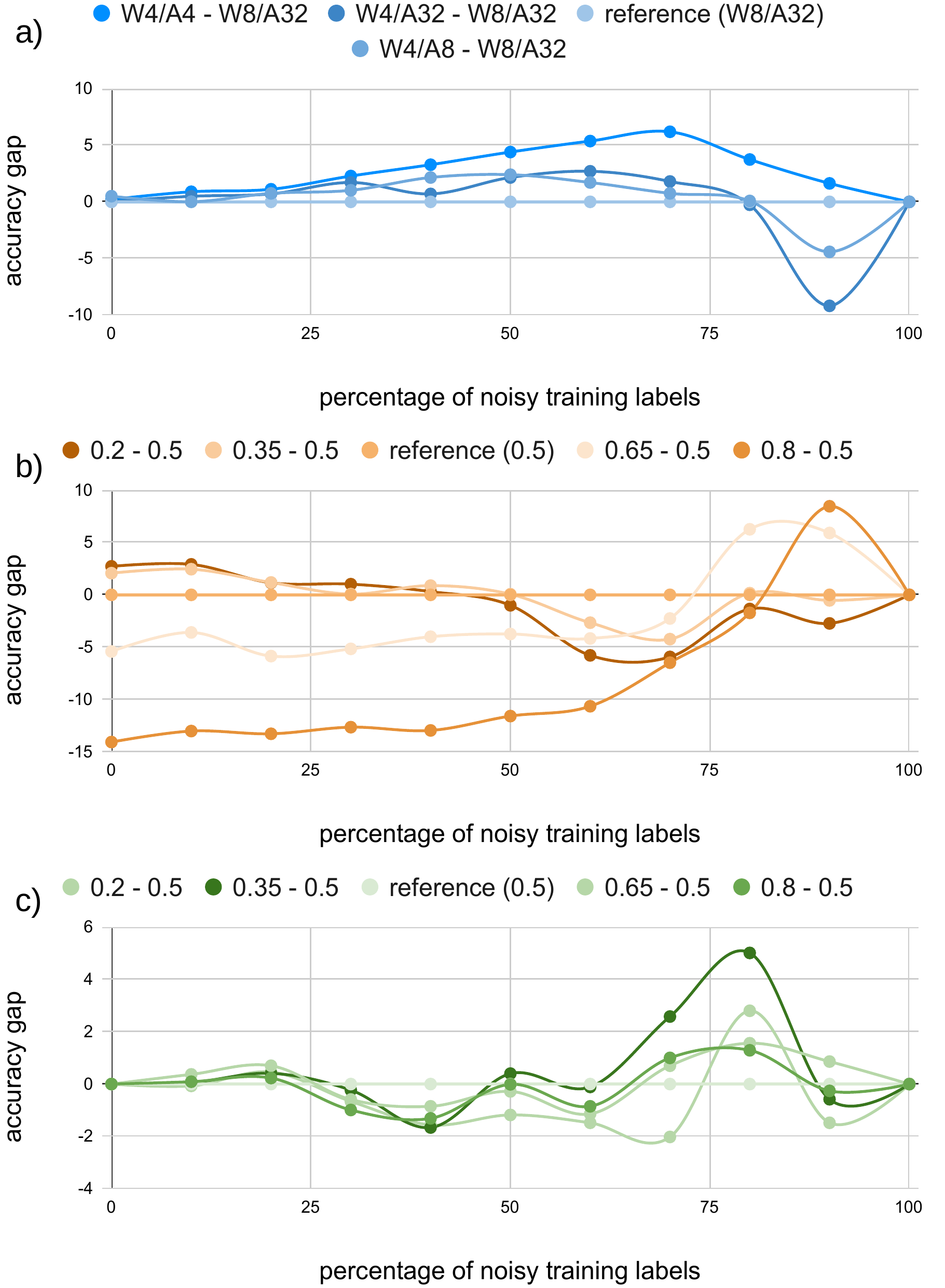}
\caption{Comparison of the accuracy gain for different hyper-parameter settings relatively to a reference value, for quantization (a), pruning (b) and dropout (c) across different values of $s$.}
\label{fig:cifar10_stability}
\vspace{-0.5cm}
\end{figure}
For pruning, it is parameterized by the proportion of neurons to remove that we set to $75\%$. Quantization is defined by the number of bits used on weights quantization and activation quantization, e.g. for int4 weights and int8 activations, we get W4/A8. As it is commonly done, we systematically quantize the first and last layers in W8/A8.

\subsection{Fighting overfitting from synthetic noisy labels}
In Fig. \ref{fig:cifar10_performance}, we compare the accuracy gain (top plot) and absolute accuracy (bottom plot) obtained from different regularization techniques as functions of the $s\%$. While early stopping and weight decay do not provide significant benefits over the baseline, we observe that quantization offers the best accuracy preservation performance for values of $s$ below $40\%$ which already correspond to almost half of the ill-annotated training examples. This result was obtained with W4/A4 quantization. This demonstrates the efficiency of quantization to tackle overfitting (observed in Fig. \ref{fig:bp4d_quantvbaseline}) from poor annotation.

For higher noise proportions, we see dropout and pruning taking the upper hand, successively. Nevertheless, quantization remains competitive; furthermore, note that these methods involve parameter tuning (e.g. for dropout the position of the dropout layer in the network as well as the dropout hyperparameter parameter), while this parameter selection is a lot coarser with quantization. To elaborate on this, we report on Fig. \ref{fig:cifar10_stability} the relative accuracy gain for different regularization techniques and parameter setups. This highlights the robustness of quantization to hyper-parameter selection; the performance of dropout and pruning, conversely, is dependant on their hyper-parameter setting, which requires validation in practice.
Consequently, quantization offers the highest accuracies in the most realistic experimental set-up (noise below $30\%$) while also offering robustness to parameter selection. 

\subsection{Fighting overfitting on BP4D}
Now that we showed on controlled experiments with synthetic label noise that quantization indeed acted as regularization, preventing overfitting on noisy labels, in what follows we demonstrate that on BP4D, a real noisy dataset \cite{zhang2014bp4d} with high inter-annotator disagreement rate, the proposed approach allows to improve the accuracy as compared with the baseline as well as existing methods to fight overfitting. 

In Table \ref{tab:regul_compare}, we observe that weight decay and label smoothing offer marginal improvements to the average F1 score at the expense of stability across AU. The best performing standard regularization technique, dropout, increases the average F1 score by 1.1 points over the baseline and 1.0 point over the second best regularization technique, weight decay. Still, quantization offers an extra 1.1 points over dropout, reduces the standard deviation across AU and, thus improves stability. Last but not least, quantization allows a significant speed-up at inference time, (up to 55\% runtime reduction \cite{jacob2018quantization}).


\section{Conclusion}

In this study, we investigated the ability of compression techniques to tackle both the challenges of preventing overfitting on noisy labels, and the difficulty to efficiently deploy trained neural networks. More specifically, in addition to existing methods geared towards preventing overfitting, such as early stopping, weight decay, label smoothing and dropout, we investigated a number of deep neural network compression techniques such as pruning and quantization. We conducted a thorough empirical validation: first on Cifar10, on which we synthetically added a proportion of erroneous labels, and, second on BP4D AU detection dataset, which is notoriously noisy due to the subtlety of the task at hand. The results show that deep neural network quantization leads to increased robustness to label noise on both simulated and real test cases, and allows superior performance as compared with other methods. Hence, quantization is an effective method for limiting overfitting on noisy data that also allows more efficient inference.

Future work involves further investigation on multi-task quantization. We observed that, despite the significant improvements, the inter-AU accuracy discrepancy in the quantized network remains high. One solution would be to design a quantization scheme that adapts the constraint for each task independently.

\bibliographystyle{IEEEbib}
\bibliography{main}

\end{document}

%% file: tables/bp4d/regularization_comparaison.tex
\resizebox{0.9\linewidth}{!}{
\begin{tabular}{|c|c|c|c|c|c|c|c|c|c|c|c|c|c|c|}
\hline
\footnotesize {\bf{F1 Score-AU}} &1 & 2 & 4 & 6 & 7 & 10 & 12 & 14 & 15 & 17 & 23 & 24 & {\bf{Avg.}} \\
\hline
\footnotesize Baseline & 49.6 & 45.8 & 55.1 & 74.8 & 75.5 & 83.7 & 87.1 & \textbf{62.7} & 47.0 & 56.4 & 44.8 & 41.2 & 60.3 \\
\footnotesize Weight Decay & \textbf{52.6} & 46.8 & 54.3 & 75.3 & 76.4  & 83.1 & 87.2 & 58.6 & 44.6 & 58.5 & 44.1 & 43.1 & 60.4 \\
\footnotesize Dropout & 50.2 & 46.1 & 57.5 & 75.3  & 75.8 & 83.1 & 87.1 & 60.5 & 47.9 & 58.4 & 46.9 & 47.9 & 61.4 \\
\footnotesize Label Smoothing  & 48.8 & 45.7 & 56.0 & 75.3 & 74.8 & 82.7 & 87.3 & 60.1 & 44.0 & 57.1 & 43.8 & 41.5 & 59.8 \\
\hline 
\footnotesize Pruning & 51.0 & 45.0 & 56.2 & \textbf{76.1} & 76.4 & 82.5 & 86.8 & 61.3 & 47.1 & 58.0 & 44.2 & 42.8 & 60.6 \\
\footnotesize Quantization & 51.7 & \textbf{48.2} & \textbf{57.6} & 75.4 & \textbf{76.9}  & \textbf{83.9} & \textbf{87.8} & 59.7 & \textbf{48.4} & \textbf{59.0} & \textbf{47.8} & \textbf{53.4} & \textbf{62.5} \\
\hline 
\end{tabular}}